\documentclass[AMA,STIX1COL]{WileyNJD-v2}

\usepackage[utf8]{inputenc}
\DeclareUnicodeCharacter{2190}{<-}

\articletype{RESEARCH ARTICLE}%

\received{}
\revised{}
\accepted{}
        
\usepackage{listings}
\usepackage{xcolor}

\usepackage{graphicx}
\usepackage{float}

\colorlet{punct}{red!60!black}
\definecolor{background}{HTML}{EEEEEE}
\definecolor{delim}{RGB}{20,105,176}
\colorlet{numb}{magenta!60!black}

\definecolor{eclipseStrings}{RGB}{42,0.0,255}
\definecolor{eclipseKeywords}{RGB}{127,0,85}
\colorlet{numb}{magenta!60!black}

\lstdefinelanguage{json}{
    basicstyle=\normalfont\ttfamily,
    commentstyle=\color{eclipseStrings}, % style of comment
    stringstyle=\color{eclipseKeywords}, % style of strings
    showstringspaces=false,
    breaklines=true,
    string=[s]{"}{"},
    comment=[l]{:\ "},
    morecomment=[l]{:"},
    literate=
        *{0}{{{\color{numb}0}}}{1}
         {1}{{{\color{numb}1}}}{1}
         {2}{{{\color{numb}2}}}{1}
         {3}{{{\color{numb}3}}}{1}
         {4}{{{\color{numb}4}}}{1}
         {5}{{{\color{numb}5}}}{1}
         {6}{{{\color{numb}6}}}{1}
         {7}{{{\color{numb}7}}}{1}
         {8}{{{\color{numb}8}}}{1}
         {9}{{{\color{numb}9}}}{1}
}

\begin{document}

\title{Providing Assurance and Scrutability on Shared Data and Machine Learning Models with Verifiable Credentials}

\author[1]{Iain Barclay*}
\author[1]{Alun Preece}
\author[1,2]{Ian Taylor}
\author[2]{Swapna K. Radha}
\author[2]{Jarek Nabrzyski}

\authormark{Barclay \textsc{et al.}}

\address[1]{\orgdiv{Crime and Security Research Institute}, \orgname{Cardiff University}, \orgaddress{\state{Cardiff}, \country{UK}}}
\address[2]{\orgdiv{Center for Research Computing}, \orgname{University of Notre Dame}, \orgaddress{\state{IN}, \country{USA}}}

\corres{*Iain Barclay, Crime and Security Research Institute, School of Computer Science and Informatics, Cardiff University, Queen's Buildings, Cardiff, CF24 3AA, UK. \email{BarclayIS@cardiff.ac.uk}}
%\presentaddress{This is sample for present address text this is sample for present address text}

\abstract[Summary]
{Adopting shared data resources requires scientists to place trust in the originators of the data. When shared data is later used in the development of artificial intelligence (AI) systems or machine learning (ML) models, the trust lineage extends to the users of the system, typically practitioners in fields such as healthcare and finance. Practitioners rely on AI developers to have used relevant, trustworthy data, but may have limited insight and recourse. This paper introduces a software architecture and implementation of a system based on design patterns from the field of self-sovereign identity. Scientists can issue signed credentials attesting to qualities of their data resources. Data contributions to ML models are recorded in a bill of materials (BOM), which is stored with the model as a verifiable credential. The BOM provides a traceable record of the supply chain for an AI system, which facilitates on-going scrutiny of the qualities of the contributing components. The verified BOM, and its linkage to certified data qualities, is used in the AI Scrutineer, a web-based tool designed to offer practitioners insight into ML model constituents and highlight any problems with adopted datasets, should they be found to have biased data or be otherwise discredited.}


\iffalse
\jnlcitation{\cname{%
\author{Barclay I.},
\author{Preece A.},
\author{Taylor I.},
\author{Radha S.} and
\author{Nabrzyski J.}
} (\cyear{2021}), 
\ctitle{Providing Assurance and Scrutability on Shared Data and Machine Learning Models with Verifiable Credentials}, \cjournal{CCPE Special Issue on Science Gateways}, \cvol{2021;00:1--15}.}
\fi

\maketitle
\section{Introduction}
In order to gain confidence in the quality and suitability of a shared dataset or machine learning (ML) model, potential users need to be able to develop and maintain trust in claims made by the originators of the asset under consideration. Thornton, Knowles and Blair\cite{thornton2021} have, for example, considered challenges faced in progressing research in the environmental science field. In seeking to maximise collaboration and adoption of shared data resources in a science gateway, Thornton and colleagues identify the pivotal role played by trust between participants in the community, and the importance participants placed on trusting the quality of the work of their peers. Specifically, relying on the work of others to make decisions that can be as significant as impacting the future of our planet ``requires trust not only in the scientific findings and resulting recommendations, but subsumed within this, trust in the people, processes, and data that led to these findings and recommendations.'' Further, they identify that the collaborative nature of environmental science means that the scientists involved need to be able to develop ``a solid basis for trusting their collaborators, the processes and data they use, and their findings, in order to make collective progress in the field.''. Historically, researchers have co-operated through informal groupings, identified in research from the early days of the World Wide Web, by Van House, et. al\cite{van1998cooperative}, as Communities of Practice (CoP) which provide and foster mutual credibility. Van House and colleagues determined that ``one way that users judge whether work is to be trusted is to look at its source: is he or she a part of our community of practice? Can he or she be trusted to have used accepted methods to collect, analyze, and interpret the data? Do we speak the same language? Do they see the world the same way that we do?''
\iffalse
In practice, even in environments where funding organizations insist on researchers sharing data, there can be a resistance to reuse available assets. Pisani et al.~\cite{pisani2018} found ``lower-than-expected reuse of shared data may be because potential secondary users have few ways of checking the quality of those data''. 
\fi

The UK's Open Data Institute (ODI) has recently published their ``Trustworthy Data Stewardship Guidebook''\cite{odi2021} which further considers the role of trust in environments looking to foster data sharing. In defining terminology, the ODI adopt ```trustworthy' to mean an organization is worthy of being trusted, while `trust' refers to an organization actually being trusted by an individual, organization or ecosystem''. As such, they argue that effective data sharing environments need to provide all participants with mechanisms to develop confidence that their fellow actors in the ecosystem are trustworthy, in order that trust can be achieved to the degree that data sharing can take place effectively. Beyond being an abstract concept, effective data sharing facilitates progress in science and ``enables organizations to operate and ecosystems to function'' through realisation of the societal and economic value of data, and associated products and services. Being demonstrably trustworthy as a data steward or a data consumer facilitates value creation whilst limiting harms\cite{odi2021}.

The potential for problems to result from a lack of visibility and an ability to assess the trustworthiness of adopted data assets increase as data is used to build new knowledge products, which move out of scientist's laboratories and into deployment environments. As domain experts and practitioners come to rely upon tools built from data and inferences made by data analysis, manifested as artificial intelligence and machine learning systems (AI/ML), there is an increasing need to provide accessible and approachable ways for these domain experts and practitioners to inspect the origins of the resources that their tools are built upon. Foreshadowing this need, USAID\cite{usaid2021} caution practitioners adopting AI/ML that ``Often technologies are developed by people and organizations who do not reflect the people and communities who are influenced by the system. Assumptions and biases can become embedded into the system, resulting in ineffective technologies or harmful consequences.'', and the UK's National Health Service explicitly points out the responsibilities of its organizations deploying AI systems by asking stakeholders the direct question: ``Can you develop a sufficiently robust understanding of relevant data feeds, flows and structures, such that if any changes occur to model data inputs, you can assess any potential impacts on model performance - or signpost questions to the vendor?''\cite{nhsX2020}. This is not an abstract concern -- in their extensive review of AI deployments in global health settings, Ismail and Kumar \cite{kumar2021} found that ``data used to develop systems was rarely exclusively generated in the region targeted'' - they learned that data was often generated from openly available sources in the US or Europe, even when the AI/ML solution was to be applied in other regions. In some cases, data was supplied by a local partner and then supplemented with data from other sources, including World Health Organization and USAID, and publicly available geospatial and meteorological data. Outside of healthcare there are publicised examples of AI systems that are discredited due to doubts being raised on the legitimacy of their data sources\cite{rezende2020facial}, or of societal bias being discovered in training data after analysis\cite{chandrabose2021overview}. Compounding this, MIT researchers\cite{northcutt2021pervasive}, have found many labelling errors in published datasets\footnote{\url{https://labelerrors.com}} that are widely used in ML model training and evaluation.

In seeking to determine how to provide improved trust and transparency between participants in shared data and knowledge-based systems, this paper looks to technologies that have been shown to provide affordances of trust between disconnected parties in different environments. Distributed ledger technologies, implemented as blockchains, offer protocols that can provide cryptographic proof that a party has written a transaction to the ledger, and mechanisms to ensure that the ledger is not subsequently altered. Building upon these foundations, self-sovereign identity (SSI) data models and protocols can further provide proof of ownership of a digital identity used to issue claims or credentials from one entity to another, such that the claims can be inspected by a third party and verified to be genuine and unchanged. Adopting these technologies and protocols offers a novel means to build trust in shared assets and overcome resistance to data sharing. To provide assurance, verifiable credentials (VC) can be used by a dataset or ML model publisher to cryptographically sign metadata, or other supplementary information pertaining to the asset and its origination. With these signed assertions in place, prospective users and other stakeholders can inspect and verify digital certificates that accompany shared ML and data assets, and be assured that the publisher is irrefutably attesting to its qualities. Providing tools that enable practitioners and domain experts to scrutinise AI and data assets, so that they are able to better understand the contributions of people and data in these assets, promises to offer better informed judgements on the ongoing suitability of data and AI/ML for use in diverse, real-world, environments.

This paper contributes to ongoing research in data and AI transparency and assurance by presenting a software architecture with the following characteristics: use of decentralized self-sovereign identity technologies to enable data owners to issue signed certifications of the qualities of their data and data scientists to inspect the qualities and gain confidence in the third party data; use of a supply chain model to catalogue the constituent components of ML models; provision of a user interface for practitioners using ML models to view the constituent parts of the model and check in real-time that contributions have not been discredited or invalidated. The architecture is evaluated through an implementation of the system, which integrates into an ML production pipeline.

The paper begins with a review of mechanisms for the provision of assurance of shared data assets, ML models and AI systems, and proposals for structures and styles of documentation for datasets, ML models and AI systems from recent literature. This review, in Section~\ref{sec:requirements}, is used to motivate discussion on the adoption of a supply chain model to represent contributions towards data and knowledge products, and define the requirements for a system that can provide assurance to users of shared datasets and ML assets. Section~\ref{sec:approach} provides an introduction to SSI technologies, which are adopted in the design of a solution architecture for a verifiable supply chain of assets in an ML model, which is discussed in Section~\ref{sec:architecture}. An implementation of the architecture is presented in Section ~\ref{sec:implementation}. The paper concludes with an evaluation of the approach and recommendations for further work in Section~\ref{sec:evaluation}.

\section{Related Work}
\label{sec:requirements}

This section begins with a review of the literature concerned with sharing of digital assets such as datasets and ML models, as well as identifying best practice for communication of the constitution of assets from established fields, including industry and software. The review further identifies research on user needs regarding adoption of shared assets, which motivates a direction for a proposed solution architecture.

\subsection{The Data Sharing Landscape}
\label{sec:barriers}
Wallis et. al\cite{wallis2013if} describe the `long tail' of research data, with small-scale projects, producing limited volumes of data that are typically only shared with trusted peers and colleagues, often as part of an informal gifting or bartering process. Niche datasets which may once have dwindled on the long tail of research in their standalone state, have an opportunity for new life and new, wider audiences when they are utilised within knowledge products. However, as these datasets are adopted and used in the creation of data-reliant products, such as ML models, sight of the original data source and knowledge of the data originators is very often lost. As such, providing robust and standardised mechanisms to document and convey the documentation about datasets is beoming increasingly important. Researchers at Google have been considering issues around dataset documentation, and in particular concerns with datasets that are subsequently used in artificial intelligence and machine learning (AI/ML) applications. Hutchinson, et. al,\cite{hutchinson2021towards} argue that data development practises are often poorly specified, and the poor-relation in regards to the time and effort that goes into developing AI algorithms. They propose that lessons are learned from software engineering, with a more rigorous approach being taken to the data development lifecycle, in order to avoid a genuine crisis in AI - asking ``how can AI systems be trusted when the processes that generate their development data are so poorly understood?''. Geiger et al \cite{geiger2020garbage}, for example, discuss a wide diversity in the level and quality of detail being provided on the labelling processes for data sourced from twitter and being used in ML model training. 

\subsection{Documenting Data for AI Systems and ML Models}
Recent years have seen research attention paid to documentation of ML models and to the data assets that contribute to the models. Gebru et al suggested `Datasheets for Datasets'\cite{gebru2018datasheets}, as an attempt to replicate the datasheet format that is common in electrical components for use with shared data assets. The Datasheet's proposal is to produce a human readable story of the data, giving data providers an opportunity to reflect on the nature of the data they are providing, and the ongoing work they need to perform to maintain that data, while providing data users -- who will typically be ML model authors -- with insights into the data's history and how the data is intended to be used. This relaying of information, almost the biography of the data, is purposeful, in that ``transparency on the part of dataset creators is necessary for dataset consumers to be sufficiently well informed that they can select appropriate datasets for their tasks and avoid unintentional misuse''. Mitchell, et al., developed this style of documentation further with a proposal for `Model Cards for Model Reporting'\cite{mitchell2019model} which aims to provide a page or two of information to accompany an ML model, and provides a framework to promote a standardisation of ``ethical practice and reporting'' to allow practitioners and stakeholders to compare models ``along the axes of ethical, inclusive, and fair considerations.''

An intention of Model Cards is to assist less technical users of ML models, and developers who wish to adopt the model as part of a wider system or workflow with insight into how the model should behave and how it can be combined with other models. In summary, Mitchell, et al. argue that researchers sharing ML models should seek to produce Model Cards to aid potential users in becoming better informed on the suitability of particular models for particular use cases.

Model Cards have been adopted by some organizations since being proposed, and to foster wider adoption, developers have provided an open source tool -- the Model Card Toolkit (MCT)\cite{MTK2020} -- and example documentation on how they can be generated and integrated into a model production workflow.

Model Cards fit within the wider context of Explainable AI\cite{bhatt2020explainable}, which seeks to unravel and present the hidden complexities of AI systems to a range of stakeholders, including other developers, and in some cases end-users. Explainability covers a wide scope, but providing information and  ``audit trails''\cite{brundage2020toward} on the parts and processes that have led to the generation of a model is recognised as a potential component, and research has shown it can be a contributor to providing confidence in adoption of AI systems among practitioners\cite{cai2019hello}, helping to bridge the ``last mile''\cite{cabitza2020bridging} of AI deployment. Furthermore, being able to clearly identify and enumerate data sources contributing to an AI asset provides a means to understand the ``bibliometric data'' behind AI systems, identified by Mishra, et al.\cite{mishra2020measurement}, as a way to assess the equity in contributions to AI research by factors including geography, gender, and other attributes.

\subsection{Documentation from Other Industries}
Outside of the AI/ML domain, organizations such as the UK's Open Data Institute\cite{odi2018} have described and illustrated efforts to map the contributors and stakeholders in situations where multiple parties cooperate in providing data sources in scenarios such as offering Transport for London's transport data\cite{stone2018improving} via an open API. These data ecosystem mappings provide a visual overview of the stakeholders in the systems, along with connections showing the data flows and value flows\cite{Attard2017} between the parties. Considering an AI system or an ML model as an output asset of such a data ecosystem, wherein multiple actors cooperate to add value to datasets, then a visual data ecosystem map could provide a suitable vehicle for providing context to the parties who have made a contribution to the AI system and be useful in augmenting textual documentation\cite{hegarty1993constructing} about the details of contributions to end-users and developers. Presenting a visual overview of an AI/ML system can help to visualise the different roles of contributors, such that documentation for each contributing entity can be understood in the context of its role in the system as a whole.

A comparison can be made to the familiar model of a physical product's supply chain, and its record of the components and sub-assemblies used in the production of the finished product in a document known as a Bill of Materials (BOM)\cite{jansen2003managing, van2003traceability}. The model of a BOM has been adopted in the software engineering community, particularly in the field of open source software, where solutions increasingly adopt libraries and components from multiple providers and the community at large. A Software Bill of Materials (SBOM) is used to document the supply chain of dependencies, so that vulnerabilities can be readily and rapidly identified\cite{dodson2019mitigating} and outdated sub-components of  fixed or replaced in updated versions of the module\cite{mackey2018building}. Tools which support SBOMs include CycloneDX\footnote{\url{https://cyclonedx.org}} and  SPDX\footnote{\url{https://spdx.org}}, and they define schemas and formats for listing and tracking sub-components. SBOMs can be generated from the build processes of software tools, as part of the development and deployment pipeline. As well as documenting dependencies, SBOM formats also provide placeholders for supplementary information such as licenses for software components and libraries used. SBOMs are typically integrated into vulnerability tracking and component analysis systems, such as Dependency Track\footnote{\url{https://dependencytrack.org}}, which provide notice of detected vulnerabilities in system sub-components, provided by sources such as publicly known vulnerabilities, such as those listed in the US Government's National Vulnerability Database\footnote{\url{https://nvd.nist.gov}}. Carmody et al\cite{carmody2021building}, describe the benefits this has brought to a medical devices system, where critical infrastructure is now protected by providing warning on vulnerabilities in previously hidden system components.

\section{Solution Approach to Providing Scrutable AI}
\label{sec:approach}
The software industry's SBOM initiative and its connections to vulnerability tracking and alerting provides a pattern upon which to develop a model for tracking the elements of an an AI/ML system, and in particular to begin to document and provide a means to trace contributions from different data sources and providers, such that any issues or vulnerabilities found in the data sources can be raised as a problem and made subject to investigation. In previous work\cite{barclay2019towards}, the authors have proposed a model of a supply chain for AI/ML products, based on the premise of ML model authors and publishers populating a BOM to detail the contributions of components, such as data sources and other artefacts, along with the human work towards creation of the resultant model.

\subsection{The AI/ML Supply Chain}
In considering the production workflows for ML models, the contributing assets can be identified and itemised, and typically include original data sources and labelled datasets used for model training and validation, along with human expertise used in data curation, and in development and testing of the model. Research on the security and integrity of ML systems identifies threat vectors which include Sybil attacks\cite{wang2016defending}, data poisoning attacks\cite{miao2018towards}, and model poisoning attacks\cite{gu2017badnets}. Further, as ML models mature and are deployed and used in production environments, it is conceivable that qualifications, best practise, and ethical or legal standards which were appropriate at development or deployment time are no longer adequate by the standards of the day. Any lack of visibility or inability to scrutinise the contributions to ML models and data assets is exacerbated as the distance between the developers and the practitioners using the model increases, as is the case when models are sourced from third parties, via commercial or community marketplace platforms\cite{zhao2018packaging}, science gateways and model zoos\cite{jia2015caffe}. USAID\cite{usaid2021} describe a situation in South Africa where the introduction of the Protection of Personal Information Act (POPA) in 2013 had serious ramifications for already deployed ML systems, as there was uncertainty about the whether data sharing and management processes were in place, as a result ``essential services, such as the distribution of medication for chronic conditions, were suspended and uncertainty about the long-term viability of using ML was introduced.''

In order to provide a supply chain for an ML model that is able to provide users and practitioners with the ability to build trust and to perform necessary and ongoing scrutiny on the model through the BOM, a robust and reliable architectural model needs to be established. This architecture needs to be able to provide those scrutinising the ML model and its contributing elements with confidence that it has been issued and endorsed by authorised and trusted parties, and that it is accurate and up-to-date. Conversely, the technology needs to be able to protect the privacy of those offering the model and contributing assets when it is necessary to do so. As such, the philosophy and technological implementation of self-sovereign identity appears to offer an interesting approach.

\subsection{Self-sovereignty of Digital Assets}
Self-sovereign Identity provides an individual with the ability to take ownership of their personal identity data and have control over access to that data, without the need for centralized infrastructure\cite{allen2016path}. With the adoption of self-sovereign identity, it is possible to secure a user's personal data from unauthorized disclosure by allowing the data owner to selectively disclose elements of their data based on the requirement from the verifying party, and the value that the data owner places on that exchange. An individual thus gains the ability to decide how identity and other personal data should be used and who has access to it. The SSI community have developed data models and protocols that provide cryptographically verifiable mechanisms for validating identities and issuing and presenting proofs of held credentials\cite{sporny2019}. These mechanisms build on decentralized technologies to provide immutable proof of control of an identifier and to facilitate the secure exchange of credentials based upon these identifiers between consenting parties. Implementations of SSI protocols can employ a blockchain ledger to provide integrity to their systems, and support further capabilities. One such implementation is the Hyperledger Aries\cite{AcaPy} platform, which uses a blockchain to secure the schema of the credentials issued, and provide an immutable record of revoked credential identifiers, such that they can’t be presented erroneously. The blockchain used by Hyperledger Aries is integral to the platform, and optimised for efficient lookup of data to support decentralized identity use cases.

Although not a primary motivation of the specification, the protocols and data models developed by researchers of SSI technologies for personal identity can be adapted and applied to other entities, including data assets and devices. One such model, and the use case presented here, is the use of a Verifiable Credential (VC) to assert claims about a digital asset, such as a dataset. This can be manifested by using a data document consisting of a set of key-value pairs to express claims made by one party about another -- the Subject of the claim. The party making the claims, known as the Issuer, can cryptographically sign the document and provide it to the Holder, typically the subject of the claims, or a representative, who stores it securely in a digital wallet. When a third party -- the Verifier -- seeks evidence of the veracity of certain facts about the Subject, they can make a request for a proof. The Holder uses elements of the VC to produce a Verifiable Presentation, which demonstrates to the Verifier that the claims are true. The use of asymmetric encryption and shared public keys in the VC protocols ensures that issued credentials are tamper-proof and that they have been signed by the identified Issuer. If the Verifier knows and trusts the identity of the Issuer, then they can make a judgement on the value they place on the claims made by the Issuer about the Subject. VCs pertaining to data assets can be used by researchers to assure themselves of the data quality, algorithms and processes used, and of the specifications of any equipment used to gather the data. They can be considered as elements of metadata signed by a project's Principal Investigator, for example, to provide assurance in the accuracy and integrity of a shared dataset. Scientists who hold the PI in high regard will be able to verify that credentials about dataset qualities were issued by the PI, and accordingly gain confidence in the quality of the data in the dataset.

\subsubsection{Using Verifiable Credentials for Assurance}
\label{usingvcs}
A core tenet of the SSI model is that parties claiming to be the controller of a DID can provide cryptographic proof that this is the case, facilitated by a protocol that provides a route to a verification mechanism. At its simplest, this proof can be provided in the form of a structured document file containing the public key of the DID, along with the methods by which a party can verify. By using the published verification mechanisms the holder of a document that purports to have been signed by the DID's controller can obtain cryptographic proof that it was indeed signed by the DID controller, and furthermore can verify that the document has not been tampered with since it was signed. 

This systematic mechanism for a party to prove that they have access to the private keys relating to a DID is utilised when issuing VCs, which can be as simple as a document claiming control of a DID. If the claim document is signed by a reputable or trusted party, and the DID of that party is known to the verifying party, then the claims in the document can be taken to have been issued by the trusted party. In other words, if a university administrator signs a document using the private key of a DID they control, then any claims in the document can be taken to be claims that the university is willing to endorse\cite{barclay2020certifying}. The credential issuer will be the trust anchor in the system, such that anyone relying on credentials provided by the issuer will need to have trust in the issuer themselves to place value on the credentials~\cite{linn2000trust}. Where the issuer is an entity such as a university, NGO or other well-regarded organization this trust may be inherent. In other circumstances the issuer may need to source credentials from bodies with a better established reputation in order to assert their own qualities as a trustworthy issuer of credentials. In other cases, where a peer-to-peer relationship can exist, between scientists, for example, identification and knowledge of the issuer as a member of a Community of Practise\cite{van1998cooperative} can be significant and powerful. Additionally, the mechanism could be used by inspectors or regulators to certify that particular guidelines or requirements have been met.
Providing a mechanism to assure verifying parties that a DID belongs to a known and trusted authority, or is a trusted peer, is considered a governance challenge, requiring both policies and infrastructure to provide a registry of some kind, maintaining trustworthy records that can be checked. One such DID scheme has been proposed which makes use of the fact that most reputable organizations run web sites, with certificates proving the legitimacy of the identity of the web site. The \textit{did:web}~\cite{Terbu2020} scheme takes advantage of this infrastructure by utilising the web site of an organization to host the DID document, resolving \textit{did:web} to a JSON-LD file located on the web site with a well known path~\cite{nottingham2010} and relying on only authorised users being able to upload files to an organization's official web site (as would generally be the case).

\noindent
\begin{minipage}{\linewidth}
\begin{lstlisting}[language=json, label={lst:uniofsci}, caption={A fragment of the UniOfScience DID Document},captionpos=b]
{
    "@context": "https://w3id.org/did/v1",
    "id": "did:web:uniofscience.com",
    "authentication": [{
        "id": "did:web:uniofscience.com",
        "type": "Ed25519VerificationKey2018",
        "controller": "did:web:uniofscience.com",
        "publicKeyBase58": "71ANMccQC..."
      }]
      ...
}
\end{lstlisting}
\end{minipage}

For direct implementation, an open source software package \textit{vc-js}~\cite{2020vcjs} can be used to generate VC documents based on DIDs using many schemes, including \textit{did:web}. To produce a VC document for an illustrative data asset, domain names were registered for \textit{UniOfScience}, a fictitious university, and to represent a web site for the dataset, at \textit{DIDdoi.com}, and DID Documents were crafted for each, such that resolving the did:web address through the published route would reach the appropriate DID Document. Listing~\ref{lst:uniofsci} shows part of the DID Document for \textit{UniOfScience}. The second requirement is a Credential Schema~\cite{sporny2019}, which defines the semantic vocabulary to be used to describe the attributes of the dataset and provides the format in which the claims about a particular subject will be made. To produce a VC document the \textit{vc-js} library can be integrated with the Node.js Express~\cite{hahn2016express} framework to enable a simple web form to be served to allow a user --- perhaps a scientist preparing to publish a dataset --- to enter information about the dataset which is subsequently used to populate data fields in the credential schema, and \textit{vc-js} invoked to encapsulate these values in a Verifiable Credential JSON-LD document containing a proof issued by the DID belonging to \textit{UniOfScience}. The inclusion of the DID of the issuer (and signatory) of the VC document enables third parties to verify its state, which is achieved by resolving the DID to locate the DID Document holding descriptors of the mechanisms for checking signatures, usually by provision of the public key. Verifiers can use methods in \textit{vc-js} to receive cryptographic proof of the authenticity of the VC document, assuring them that it hasn't been tampered with since it was issued. As the payload of the VC document contains the DID of the dataset that it refers to, verifiers have cryptographic proof that the issuer has signed a document attesting to the properties of the DID of the subject. If the DID relates to a dataset, then the verifier can be assured that the VC document carries signed assurances about the properties of the dataset.

\iffalse
Presentation of a VC document in this scheme provides systematic improvements over an adhoc digitally signed document, through the publication of the location of the public key of the subjects and the use of a semantic vocabulary for expressing metadata claims, but still exhibits shortcomings. Among these is the possibility for anyone who holds a copy of the VC document to present it, even though it may no longer be relevant or valid. If this is done with ill-intent it could be considered a replay attack~\cite{nist2017}. A mechanism to prevent such replay attacks is for the verifier of a credential claim to ask the holder to present a Verifiable Proof, which takes the form of a JSON-LD document signed by the VC holder containing a challenge, typically a nonce, issued by the verifier along with the credential and its proof. By inspection of the Verifiable Proof document through resolution of the DID of its issuer, and comparing this DID with the DID of the VC's subject, the verifier can determine that the challenge response is acceptable, and that the holder of the VC is still content to present it. Further verification of the credential contained in the Verifiable Proof can demonstrate that the credential has not been tampered with, and thereby provide assurance that a valid credential about the dataset has been issued by an authorised party, to a holder who still considers the credential appropriate to share. 
\fi

The process described thus far has implied that parties wishing to verify credentials make requests and human operators are on hand to manage private keys and to create and sign Verifiable Proof documents, which would soon become impractical where there was high demand or a need for a timely response. By considering a digital asset such as dataset or an ML model as a self-sovereign entity in its own right, with control over its own credentials, it is possible to begin to automate these processes. The open source Hyperledger Aries project~\cite{Aries}, provides an environment to support this paradigm, whereby any entity (e.g. an ML model) can be represented by a software application (termed an agent) operating on behalf of the entity and mediating access to the entity's credentials. 

In a data sharing scenario, a cloud-based software agent (DSA) would represent the published dataset (DS) and provide mechanisms to generate and store the private keys and issued VC documents securely in a digital wallet. Parties with interests in the dataset will interact with the representing agent by addressing it through its publicly shared decentralized identifier (DID), which is facilitated by the Aries platform. The publisher of the dataset will also have a software agent representing their role in the transactions, with the capability to issue credentials. The publisher's agent will be the trust anchor in the system, such that anyone relying on data credentials provided by the publisher's agent will need to have trust in the publisher themselves in order to place value on the credentials. In the first instance the university will configure and launch a new software process to act as the agent DSA and represent DS in future transactions. This software process will be part of the infrastructure provided by an implementation of the SSI protocols, such as the ACA-Py~\cite{AcaPy} cloud agent component from the Hyperledger Aries project. As part of the on-boarding procedure for the agent, a configuration script will be run to generate a digital wallet for the agent, which will hold its private keys and credentials, and generate a DID and an endpoint, by which DSA will be addressed and accessed.

\begin{figure} [th]
\centering
\includegraphics[width=0.45\textwidth]{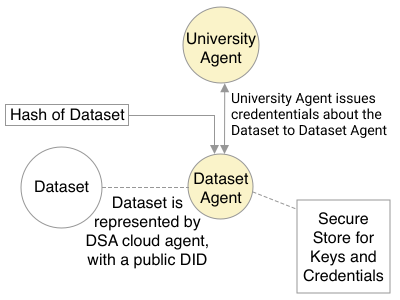} 
\caption{A publisher agent (University Agent) interacts and issues credentials to a dataset agent.}
\label{fig:did}
\end{figure}

The university will then establish a secure connection between their publisher agent and DSA, which will be manifested using Aries' SSI protocols, and the resultant connection will be used to issue the appropriate credentials to DSA, as shown in Figure~\ref{fig:did}. Credentials are structured according to published JSON-formatted Credential Definition schema~\cite{lux2019full} and contain a set of key-value pairs which the issuing party asserts are true about the holding entity. For example, in Listing~\ref{lst:hash}, \textit{Hash of Data} is a credential holding the cryptographic hash of the dataset, which inextricably links the credential to the dataset it represents, and \textit{Data Ethically Sourced} represents an ethical status of the dataset, as stated by the publisher. A practical scheme will hold other credentials, and could include a credential expiry date or other conditions for credential usage. 

\noindent
\begin{minipage}{\linewidth}
\begin{lstlisting}[language=json, label={lst:hash}, caption={A sample credential set},captionpos=b]
{
    "Hash of Data": "0xFFEE...AA1122",
    "Data Ethically Sourced": "YES"
}
\end{lstlisting}
\end{minipage}

\begin{figure} [th]
\centering
\includegraphics[width=0.45\textwidth]{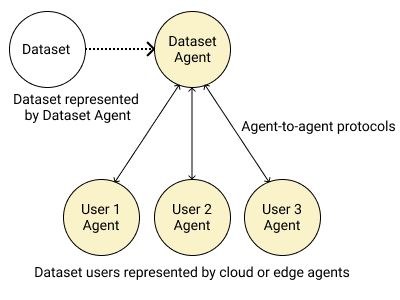}
\caption{Agents represent parties in the relationship.}
\label{fig:data_agents}
\end{figure}

The DID of the dataset agent (DSA) is a public address, analogous to a website address, which can be shared in a downloadable package for the dataset (DS) that it represents, such that users can use this address to request proof of the credentials of DS. Users of DS will themselves use software agents to communicate with its agent, DSA.  These may be edge agents stored on a mobile device or cloud agents hosted by a web service. Any user with knowledge of the DID for DSA can seek to establish a connection with DSA and to request proof of the DS's credentials via their own agent, and if the system policies permit, DSA will respond, presenting proof of the credentials it holds. These credentials will include the cryptographic hash of DS, to provide an inextricable link to the underlying dataset it represents, along with its ethical sourcing status, as written in the credential by the publisher. The public DID of the university will be included in the returned credential proof, and can be used as a trust anchor to assure that the credentials originated from the university. DID protocols ensure that the returned proof is cryptographically provable to have originated from the issuing university and to have not expired, been revoked or tampered with in any way. The architecture for such a scheme is shown in Figure~\ref{fig:data_agents}.

In an extension of this scenario, an additional cloud-based software agent could represent a published ML model. This ML agent could hold a BOM document describing the components that had been used in its development, which in turn would reference datasets used in the model's development. The BOM document could be issued as a VC by the ML model owner or publisher, and could be held securely by a software agent representing the ML model. Any party interested in the model could access the BOM through the model's agent, and receive a verified document detailing the model's components. Design and development of such a system and its use to provide insight into ML model constituents is described in subsequent sections.

\section{Architecture and Design}
\label{sec:architecture}

To demonstrate and evaluate the use of VCs to provide verifiable BOM information for an ML model a system, entitled The AI Scrutineer (AIS), was envisaged. The AIS system provides a means for the publisher of an ML Model to create a BOM document for the model, such that parties that are reliant on the ML model can inspect its on-going suitability for use. AIS is designed to be readily accessible by a practitioner using an ML model, and as such it provides a web-based user interface that offers the model's user insight into the make-up of a model.

\begin{figure} [th]
\centering
\includegraphics[width=0.8\textwidth]{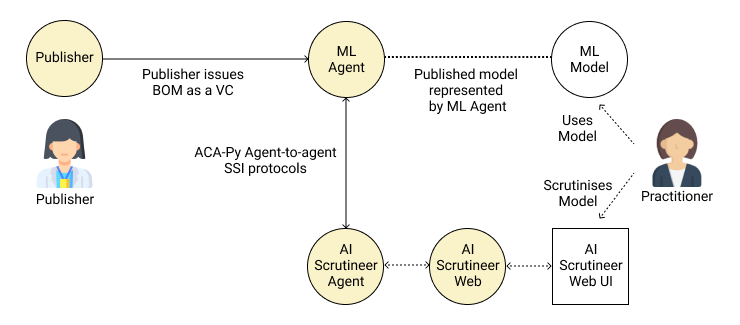} 
\caption{Conceptual Architecture of The AI Scrutineer}
\label{fig:agents}
\end{figure}

There are three different roles in the system, each represented by an autonomous software agent: there is an agent which represents the ML model's publisher, this agent is considered to be the root of trust in the system, and is the party that issues credentials to the second agent in the system, that which represents the model itself. The third agent is the verifying party - the agent that represents an actor who wishes to make enquiries about the ML model - it is envisaged that this would be a domain practitioner, or end-user of the model. Note that an assumption is made that the verifying party (the practitioner, in this case) trusts the ML publisher to provide accurate information about the model, such that they are comfortable to have confidence in the veracity of any information provided to them by the publisher. The AIS system (by virtue of adoption of SSI protocols) provides mechanisms that can cryptographically prove that information has come from the publisher, and has not been tampered with, but the utility of this information is dependant on the verifying party having confidence that the credential issuer is trustworthy.

In order that the AIS system can be used by practitioners in the field, who are unlikely to have knowledge of SSI or access to the software required to request credentials from other systems, a web interface is provided to enable non-experts to interact with an agent representing the verifying party. As such, a practitioner can make a request to a web page which can invoke a series of interactions on their behalf, and present the information in a meaningful and accessible manner. The intent of the system is to provide practitioners a means to access verifiable, up-to-date information on the status of contributions to an ML model or AI system, and that is manifested through the web interface, which can be accessed through a published URL or conveniently accessed by scanning a QR code with the URL embedded. The conceptual architecture of the AIS system is shown in Figure~\ref{fig:agents}.

AIS seeks to provide a verifiable BOM that describes the contributions to an ML model, such that practitioners can be assured that the model has ongoing suitability for their use. As such, it seeks to provide information on the model itself, the datasets used to develop the model, and other artifacts that contributed towards the generation of the model, along with any information that can be provided about the human contributors to the model.

When a model is being prepared to be published and shared its BOM is populated by its publisher. The BOM structure is defined by a JSON schema (described below) which provides a set of key-value pairs in which the publisher can provide relevant metadata about the model. This metadata can include information on datasets used to train and evaluate the model, as well as any related artefacts that contributed to the model's creation. Information on human contributors to the model can also be provided. Populating this information about the model will be part of the production pipeline for publishing ML models, and the model publisher's goal should be to provide sufficient and accurate information to allow remote practitioners to use the BOM to assess the virtues of the model.

The JSON structure for the BOM is encoded as a VC and signed by the publisher and issued to the software agent representing the ML model. The BOM is intended to be machine readable, such that it can be processed automatically and rendered into a user interface for presentation to practitioners. To provide full traceability through the BOM, it is intended that contributing datasets and human contributors are referenced by DIDs. In this way, the AIS process can traverse the hierarchy of the BOM for a given model and request verifiable credential assurances for all the contributing elements listed in the BOM. This will enable the AIS to provide a user interface to the practitioner which provides real-time proof of the credentials of the model and its contributions. If at any time after publication of a model it is discovered that the model itself, or a contributing dataset or human contributor, is no longer fit for purpose or as suitable as originally thought, then its credentials can be revoked. When the AIS inspects the BOM of the model, this new state will be reflected in the information returned, and the AIS will be able to alert the practitioner of the possibility of a problem that requires further investigation.

\section{Implementation}
\label{sec:implementation}

An implementation of the AIS system was undertaken to validate and gain further insight into the architectural approach. The Hyperledger Aries SSI platform\cite{Aries} was adopted as the underlying infrastructure, to provide implementation of SSI protocols and data models, and software agents to represent entities in the system. The AIS system is built upon ACA-Py~\cite{AcaPy}, a framework from within the Hyperledger Aries project which provides software infrastructure for a network of cloud-based SSI agents, which are capable of interacting to create, issue and securely store credentials. ACA-Py provides a means for application developers to implement service and business logic for SSI implementations at a higher level of abstraction. The research team further developed a framework, Syndicate.id\cite{barclayforthcoming}, on top of the Aries platform, in order to provide a higher level of abstraction to support efficient implementation of the entities and interactions of the use case, and to support future work in applying SSI models in machine-to-machine environments. Syndicate.id, is developed in the Go programming language, and provides an API to enable researchers to interact with the underlying Aries platform via http interfaces from application code.

To be most effective, the AIS implementation sought to take advantage of any documentation that already exists, so as to not place additional administrative burden on researchers and publishers of ML models. As such, the AIS BOM document adopts some elements of the Model Cards for Model Reporting \cite{mitchell2019model} documentation format to provide insight into the ML model. As part of the system design, AIS was developed to integrate with The Model Card Toolkit (MCT)\cite{MTK2020}, a software tool published by Google researchers in support of Model Cards for Model Reporting\cite{mitchell2019model}. MCT is an open source program which can be included into the production workflow of TensorFlow\footnote{url{https://www.tensorflow.org/}} model development and deployment. MCT interfaces with TensorFlow deployment tools, and is able to extract metadata from ML models. The AIS implementation has been designed such that some of this metadata relating to the model's current version is used to populate the BOM for the model.

The AI Scrutineer system is developed in both Python and Go, to facilitate integration with the Model Card Toolkit and the Syndicate.id framework which provides an interface to the SSI infrastructure provided by Hyperledger Aries and ACA-Py. Interaction between the agents representing each component of the system is conducted via http requests made to processes listening on unique ports on the host computer. The system comprises an agent process for the Publisher, an agent process for the ML model, and an agent process for the Scrutineer. Additional agents can be instantiated to represent datasets, or human operators, which is discussed below.

\subsection{Publisher Agent}
The Publisher is the party responsible for populating the BOM and signing and issuing the VC that contains the BOM data to the ML agent. In the implementation of the AIS system, a Jupyter notebook is used to interface with the publisher agent. The notebook has integration with the MCT and is able to dynamically pull information from the TensorFlow production pipeline about the model to populate the BOM. In particular, the MCT provides the name, overview and version number of the model. Other fields of the BOM are populated manually in the Jupyter notebook, these provide information about the datasets used to train the model, and brief biographical information about the data scientists responsible for production of the model.

On completion of population of the BOM data structure in the Jupyter notebook, the data is sent to the publisher agent via an http request. The handling code for the http request creates a connection between the publisher agent and the agent representing the ML model. The implementation of this connection is provided by the Hyperledger Aries platform, abstracted through Syndicate.id. AIS accesses this functionality through a Golang code interface, which enables developers to design business logic which utilises the underlying SSI protocols and data models. Once the connection between the two agents is made, the BOM data is encoded as the payload of a VC. The VC is cryptographically signed by the publisher's agent, and is issued to the ML model's agent.

The publisher's agent does not need to remain running and be accessible. Its role is to ingest the BOM data, and create, sign and issue the VC to the ML model agent and then it can terminate. The DID of the publisher agent underpins the trust in the BOM as a credential. Any party making an inspection of the BOM will rely on its veracity as a result of it being signed by the publisher, based on the publisher's DID.

\subsection{ML Model Agent}
\begin{figure} 
\centering
\includegraphics[width=0.8\textwidth]{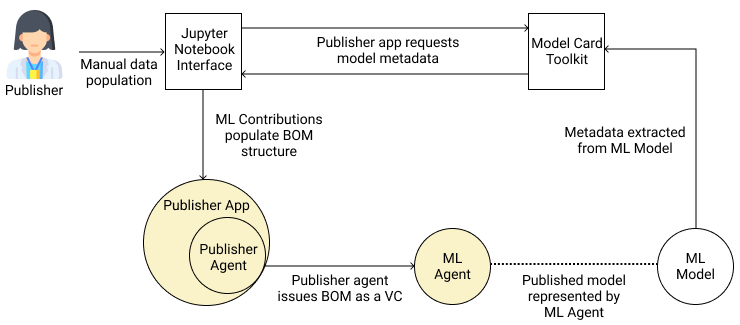} 
\caption{Interaction between Publisher Agent and Model}
\label{fig:interaction}
\end{figure}

The ML Model agent is a software agent which represents the shared ML Model. This agent will hold the BOM VC issued by the publisher, and interact with any other party which wishes to request information about the model. The model agent needs to remain running at all times, and needs to be accessible at a known endpoint location such that parties can request credentials from it. In the AIS system the model agent is a simple instantiation of an Aries ACA-Py agent, provided through the Syndicate.id framework, and doesn't have custom functionality. Its task is simply to hold a credential and to provide a presentation of that credential on request. Figure \ref{fig:interaction} shows the relationship between the Publisher and Model agents and the supporting infrastructure which populates and issues a BOM VC for a model.

\subsection{Scrutineer Agent and Web Interface}
Any party with knowledge of the endpoint address or location of the model agent can request its BOM VC, and if the request is met can inspect the BOM contents. As the AIS is intended to provide a way for non-experts to inspect ML models, the Scrutineer Agent has been developed to perform this role. The Scrutineer in the AIS implementation represents the practitioner, or any party who wishes to learn more about the ML model. 

When a practitioner wants to inspect an ML model, they visit the AIS web page, either by entering a URL or by scanning a QR code with the URL embedded, which might be provided by model publishers or in model respositories. On receiving the user's request, the Scrutineer agent makes a connection to the ML agent and asks it to provide a verifiable presentation of its BOM credential. On delivery of the credential, the Scrutineer agent decodes the underlying JSON payload data from the BOM credential and offers this in a human-readable form to the practitioner in a web browser based UI. By adopting SSI protocols, the information presented to the practitioner is cryptographically proven to have been provided by the publisher in a VC, and not altered since it was issued. If any data or other contributions are subsequently been found to be unsuitable, the VC can be reissued by the publisher to reflect this fact, or the VC could be revoked, which would be reflected in the Scrutineer's user interface.

\subsubsection{Dataset and Human Agents}
Further to the discussion above, a similar type of agent to the model agent can be provided and maintained for datasets that contribute to the model. If the BOM VC contains the DID of a dataset, then the dataset (via its agent) will be intrinsically linked to to the model. The Scrutineer agent will be able to request credentials from datasets listed in the model's BOM and show verified and up-to-date information about the datasets. Indeed, it is plausible that a dataset could be flagged as unsuitable before the model metadata was updated (and in some cases, the model information may never be updated) - as such providing DIDs of datasets in ML model BOMs provides the very best source of accurate information to practitioners.

More controversially, DIDs could be provided for human contributors, such as programmers, or data scientists. In this way, the scrutineer agent could request proof of the human contributor's qualifications. There are clearly personal privacy concerns to understand and address before it is considered desirable to include human DIDs in the BOM in this way.

\begin{figure}
\centering
\includegraphics[width=0.65\textwidth]{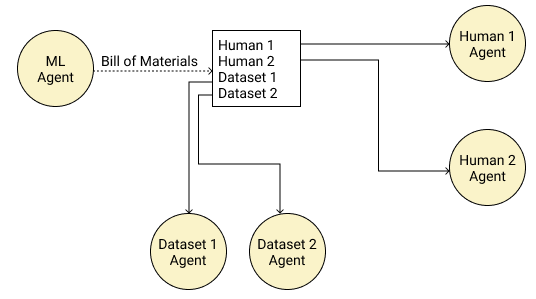} 
\caption{Linking a Model BOM to Datasets and Human Contributors}
\label{fig:hierarchy}
\end{figure}

Figure~\ref{fig:hierarchy} shows how agents representing data sources and human contributors would link to the model via the BOM held by the model's agent.

\subsection{AI Scrutineer Experimentation}
In order to exercise the AI Scrutineer system design an environment was established in which an integration with the MCT could be achieved as part of the workflow of the TensorFlow Extended (TFX)\footnote{url{https://www.tensorflow.org/tfx}} ML production pipeline. The AIS setup was integrated with a demonstration of MCT as part of a TFX example provided by researchers at Google\footnote{\url{https://www.tensorflow.org/responsible_ai/model_card_toolkit/examples/MLMD_Model_Card_Toolkit_Demo}}. The chosen demonstration integrated with TensorFlow, but alternative integrations are also provided for Scikit-learn, and the MCT can also be used as a stand-alone tool. The provided demonstration code was extended such that values from the MCT instantiation process, as well as simulated metadata representing other aspects of the data and human contributions to the example model were used to populate a BOM structure for the model, and provided to the Publisher agent via an http interface. The Publisher agent used the transferred BOM to populate a verifiable credential, which was signed and issued to the ML agent which represented the underlying model.

\begin{figure}
\centering
\includegraphics[width=0.95\textwidth]{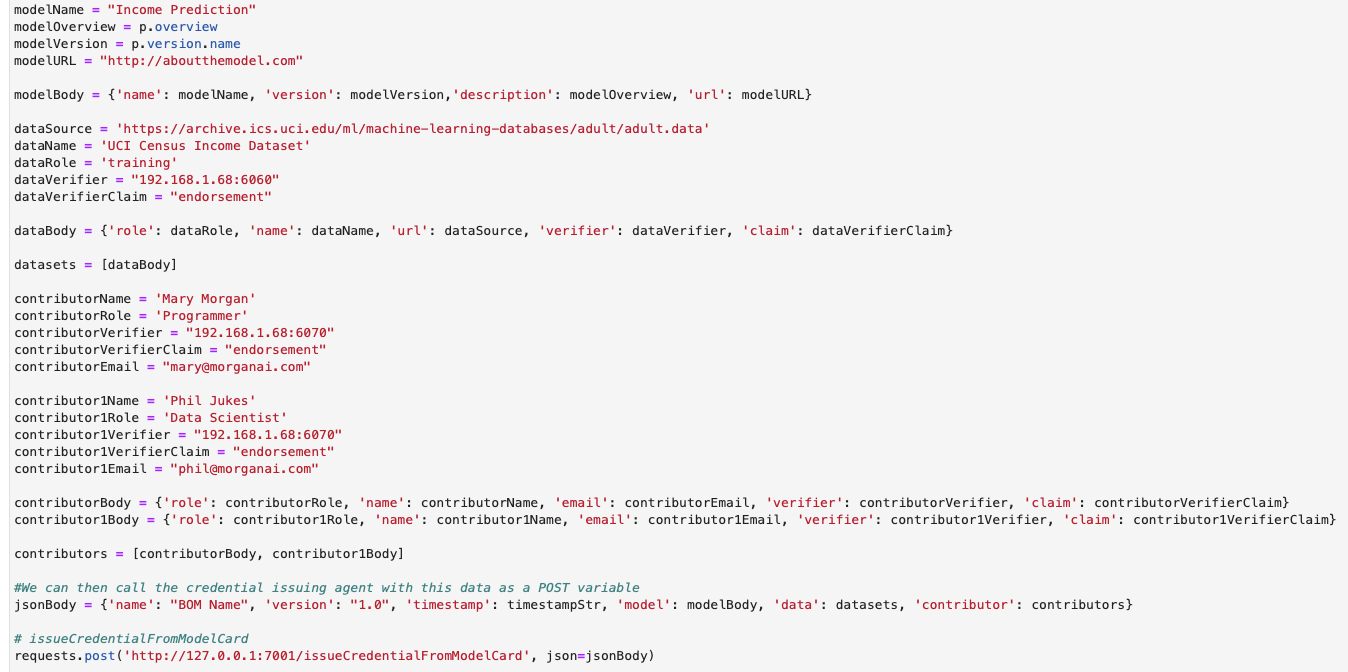}
\caption{The TFX Production Pipeline populates the BOM for the Model}
\label{fig:scrutineerpipe}
\end{figure}

After consideration, the published JSON schema for SBOM documents (used to track software dependancies) do not provide an ideal match for the requirements of an ML Model BOM, and so a very simple new schema was developed for this implementation. The BOM schema (Listing~\ref{lst:schema}) currently allows contributing datasets and human contributors to be listed, along with some general metadata about the model and the BOM itself. It is envisaged that this schema will be extended as further work is done to analyse requirements. Figure \ref{fig:scrutineerpipe} shows this schema being populated in the Jupyter notebook for the example model as part of the TensorFlow production pipeline.

\noindent
\begin{minipage}{\linewidth}
\begin{lstlisting}[language=json, label={lst:schema}, caption={Schema for the ML Model BOM},captionpos=b]
{
    "name":"BOM Name",
    "timestamp":"27-Apr-2021 10:22:52",
    "version":"1.0",

    "model":
    {
        "description": "Model description",
        "name": "Model name",
        "url": "http://aboutthemodel.com",
        "version": "1.0"
    },
    "data" : [
        {
            "name":"Dataset Name",
            "role":"training",
            "url":"https://dataurl.com",
            "verifier":"192.168.1.68:6060",
            "claim":"endorsement"
        }
    ],
    "contributor": [
        {
            "email":"mary@morganai.com",
            "name":"Mary Morgan",
		    "role":"scientist",
            "verifier":"192.168.1.68:6070",
            "claim":"endorsement"
         }
    ]
}

\end{lstlisting}
\end{minipage}

\begin{figure}
\centering
\fbox{\includegraphics[width=0.95\textwidth]{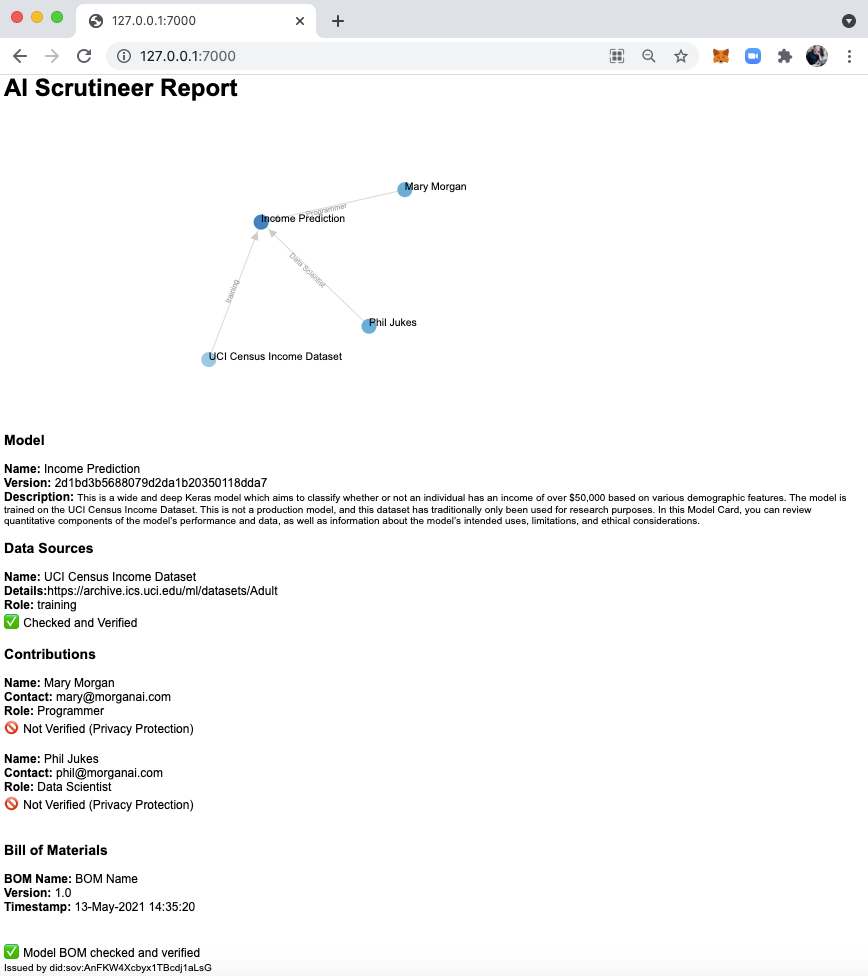}}
\caption{Screenshot of the AI Scrutineer web interface}
\label{fig:scrutineerweb}
\end{figure}

The system provided for a practitioner to inspect the model through a web interface, such that the details of the integration between the underlying SSI agents in the system were hidden, and only relevant information was displayed. Figure \ref{fig:scrutineerweb} shows a screenshot of an early version of the AI Scrutineer web interface. The user interface shows a simple graph view of the supply chain for the model, with the dataset and human contributions enumerated and linked to the model. Further details on the model (which were taken from the TensorFlow model itself, and encoded in the VC) are shown below the graph, along with some biographical information about the human contributors and the training dataset used for the developing the model. The green tick mark illustrates that the information displayed has been retrieved from a VC that has been signed by the publisher, and has been cryptographically proven (via SSI protocols in ACA-Py) to be valid and intact from the time of signature. The outcome of this approach is that if the practitioner using the AI Scrutineer system trusts the publisher of the ML model, then they can be assured that the information displayed is accurate and up-to-date.

\section{Evaluation of the Solution Approach}
\label{sec:evaluation}
This paper has presented a software architecture which adopts patterns from SSI research to enable data owners to attest to the qualities and attributes of data they wish to share, in order to demonstrate to data scientists seeking to adopt data that it can be considered to be of good quality, based on the reputation of the researchers or publishers signing and issuing the digital credential. The use of data models and protocols such as DIDs and VCs provides a cryptographically robust means of demonstrating that data can be regarded as credible, which has been identified in prior research as being one of the key criteria for the adoption of data from other sources. In effect, the DID and VCs issued to a dataset can express membership of a Community of Practice, as described by Van House, et. al.\cite{van1998cooperative}

The architecture builds upon the SSI foundations that attest to shared data qualities by providing a means for researchers and data scientists developing AI/ML models to enumerate datasets used in generation or evaluation of the model in a BOM document which describes the supply chain for the asset. The BOM itself can be signed by an AI Engineer responsible for the model, and held by a software agent which represents the model.

The BOM can be requested and inspected by model users who need to be assured of the integrity of the constituents of the model. The architecture proposes that a web-based interface is provided for domain practitioners to inspect the model, such that they can evaluate the suitability of its underlying data for their use case and domain. As such, the system is designed to provide transparency to enlighten those on the ``last mile''\cite{cabitza2020bridging} of AI deployments, as a component of offering assurance in the AI system.

The architecture has been enumerated with an implementation, which has provided a mechanism for a practitioner user of an AI/ML system to gain insight into the constituent components of an ML model by accessing a web interface which offers an overview of the model. The web interface is dynamically populated from information retrieved from the BOM, held by a software agent representing the model, and the information displayed is backed by a visual mark of assurance that it has been cryptographically verified. This verification is an important aspect of the approach, as it shows that the information has not been altered or revoked, and as a result can be taken to be as accurate as possible. As such, practitioners can gain confidence in the documentation provided with their ML model.

Design of the architecture and the implementation of the AIS system has demonstrated that it is technically viable to use SSI DIDs and VCs to attest data qualities. An ML model production pipeline has been extended to integrate with processes that populate a BOM, cryptographically signs the BOM, and issues it to a software agent. The implementation of the system has demonstrated that VC’s of datasets and ML models can be accessed and verified, and interpreted for a display to an non-expert end user, or domain practitioner.

Practically, there are still challenges to overcome in terms of operationalizing the deployment of the system, including underlying Hyperledger Aries framework, and in provisioning and deploying a set of software agents to represent each entity of the system. This will require further work to understand how a system can be deployed and how it can operate at scale. Further analysis and development of a use case scenario will enable this to be understood and mitigated. It should be noted that the AIS system is developed on SSI software which follows standardised data models and protocols, and so the design principles and interactions can be ported to other SSI platforms as they become production ready, which may provide a robust foundation for further work.

\section{Future Work}
The AIS system has been presented as an architecture of an SSI-based approach to providing a verified supply chain for an ML model and its contributions, along with an implementation that illustrates how a practitioner or user of the AI system might access documentation and gain assurance about qualities of the model. A future step in this work is to implement documentation for production models as part of the ML deployment pipeline, and to perform an external evaluation of the approach with AI engineers, end users and policy makers in order to refine the system to best effect and to offer maximum benefit to all stakeholders. If the system shows that it has utility, then further work can be conducted to develop and operationalize the approach such that it is suitable for deployment.

When the constituents of an asset are identifiable through documentation of their supply chain, a visibility metric can be determined for the asset as a whole. One such metric for ML Models (developed in our previous work\cite{barclay2020framework}) is based on adaptation of an industrial supply chain visibility metric, and considers visibility afforded to individual components of an AI/ML model through its supply chain documentation. The metric is derived for an ML model from a framework which considers the documentation supplied by the publishers of the model, and rates each element of the documentation for factors including freshness, completeness and accuracy on a likert scale. The rankings for each contribution to the model are combined to give a numeric rating for the overall visibility of the contributions to the model, representative of the visibility that those outside the development team have on the constituent components of the model. Whilst the visibility metric is a subjective ranking, it can be used across models to provide a comparative ranking of the visibility into each, or as a guide to improve documentation. Providing and displaying a cryptographically ratified supply chain of a digital asset's constituents, generated on demand, is intended to facilitate provision of good visibility on the asset and to lead to a strong score on the visibility metric. Future work is proposed to conduct an evaluation of the AIS system and the metric with practitioner users to determine the extent to which it is found to be valuable.

There are many more avenues in which to extend the work. The AIS system has provided an instance of the Scrutineer agent operating as a singular entity, yet the decentralized nature of SSI means that many different parties can operate Scrutineer agents and present the model's BOM information in different ways, appropriate to their audience - the only information that needs to be known to the scrutineer agent is the endpoint of the agent representing the model, and the credential which needs to be requested. That is not to say, however, that all data will be freely made available to all scrutineer agents. The agent receiving a credential request (in this case, the model agent) can determine whether it wishes to respond to the request, and further can decide what to return to the request, through a model of selective disclosure which protects the privacy of the actors in the system. As such, different levels of transparency -- or opacity -- could be provided for different scrutineers, providing different opportunities for information provision to partners and customers or the wider public, for example. Most significantly, the information that should be made available on human contributors to an ML system or a dataset, and how that information is shared with third parties needs very careful research as to what can usefully be provided without infringing any personal data or privacy rights of the individuals concerned.

A further area of research interest would be to consider how the information should be presented to practitioners to best effect. This would involve collaboration with human-computer interface experts, in order to understand how this complex, live, hierarchical information can be expressed most clearly and effectively. Insight may be gained from the open source software domain, where BOMs are used to alert users of vulnerabilities in underlying software components, although this audience is typically other software developers, and not end-user domain experts or practitioners.

\section{Conclusions}
\label{sec:ai_conclusion}

The architecture and implementation of the AI Scrutineer system described in this paper has demonstrated that SSI data models and protocols in the form of DIDs and VCs can be used to assert and ratify the properties and qualities of a dataset. The solution has demonstrated that a university or other data publisher can issue credentials that provide information on the metadata or other qualities of their datasets, which can be securely held by software agents, and provided on-demand to show the claims made by the publisher. If circumstances change, the claims can be revoked, and parties inspecting the dataset will be able to determine that the dataset is no longer suitable. Beyond ratifying single datasets, DIDs and VCs can be used to manifest a secure and immutable supply chain BOM of the contributions to an ML model or AI system. This can be used as part of a facility to provide assurance of the system's ongoing integrity to practitioners using the ML model, who can access it directly or through a web-based users interface as demonstrated in the AI Scrutineer implementation. Such a system can protect practitioners from using AI and ML systems where the original training or test datasets have been discredited, as real-time integrity checks on the supply chain of the asset can be performed and presented.

Whilst work remains to be done in regards to researching effective user interface design for presenting information to end users, as well as further integrating the system into data and ML production pipelines and operationalizing for deployment, the technical approach has been shown to be effective in meeting its design goals. The architecture and implementation has shown that SSI protocols and data models can be used to add assurance to data and AI systems, and provide mechanisms that can build and maintain trust between the different actors in the system, which will lead to more effective data sharing and better visibility into ML systems. 

\section*{Acknowledgments}
This research was sponsored by the U.S. Army Research Laboratory and the UK Ministry of Defence under Agreement Number W911NF-16-3-0001. The views and conclusions contained in this document are those of the authors and should not be interpreted as representing the official policies, either expressed or implied, of the U.S. Army Research Laboratory, the U.S. Government, the UK Ministry of Defence or the UK Government. The U.S. and UK Governments are authorized to reproduce and distribute reprints for Government purposes notwithstanding any copyright notation hereon.



%\balance


\bibliography{datasharing}

\end{document}